# Generating Bayesian Networks from Probability Logic Knowledge Bases

**Peter Haddawy**
Department of Electrical Engineering and Computer Science
University of Wisconsin-Milwaukee
PO Box 784
Milwaukee, WI 53201
*haddawy@cs.uwm.edu*

## Abstract

We present a method for dynamically generating Bayesian networks from knowledge bases consisting of first-order probability logic sentences. We present a subset of probability logic sufficient for representing the class of Bayesian networks with discrete-valued nodes. We impose constraints on the form of the sentences that guarantee that the knowledge base contains all the probabilistic information necessary to generate a network. We define the concept of d-separation for knowledge bases and prove that a knowledge base with independence conditions defined by d-separation is a complete specification of a probability distribution. We present a network generation algorithm that, given an inference problem in the form of a query $Q$ and a set of evidence $E$, generates a network to compute $P(Q|E)$. We prove the algorithm to be correct.

## 1 Introduction

The flexibility of Bayesian networks for representing probabilistic dependencies and the relative efficiency of computational techniques for performing inference over them makes Bayesian networks an extremely powerful tool for solving problems involving uncertainty. But Bayesian networks have two limitations that tend to restrict their use to modeling relatively narrow domains. First, they are basically propositional, i.e., nodes in a network represent multi-valued random variables. Thus to describe a class of actions or a class of individuals, one must represent each action or individual with a separate proposition. Second, a Bayesian network is a static representation in which the entire domain model is used each time an inference is performed, even though only a small portion of the network may be relevant to a particular inference. Since the general problem of inference in Bayesian networks is NP-hard [Cooper, 1990], this feature limits the size of domains that can be effectively modeled.

The approach known as knowledge-based model construction [Wellman *et al.*, 1992] has attempted to address these two limitations. The idea is to represent a class of networks with a knowledge base of probability sentences containing quantified variables and to instantiate a subset of these sentences to generate a network for solving a given inference problem. The generated network is a subset of the domain model represented by the collection of sentences in the knowledge base. Several concrete approaches to achieving this functionality have been proposed [Poole, 1993, Goldman and Charniak, 1990, Goldman and Charniak, 1993, Breese, 1992, Bacchus, 1993] But none of the approaches that provides a network generation algorithm presents a complete semantics for the knowledge base representation language, independent of that particular generation algorithm. Such a semantics is important for two reasons. First, since a user of a network generation system encodes information in the knowledge representation language, he must know the precise meanings of sentences in that language. As Wellman, et.al. [1992] point out, this semantics should be specified without recourse to the details of the network generation process. Second, one would like to be able to prove that the network generation algorithm is correct. In order to be correct, the algorithm must faithfully map probabilistic relations in the knowledge base into probabilistic relations in the network. In order to be able to prove this, we must have a formal specification of the probabilistic relations in the knowledge base.

We address the problem of providing a formal semantics for the knowledge base by using first-order probability logic as the representation language. We present a subset of probability logic sufficient for representing the class of Bayesian networks with discrete-valued nodes. We impose constraints on the form of the sentences that guarantee that the knowledge base contains all the probabilistic information necessary to generate a network. We define an independence semantics for the knowledge base that is analogous to Pearl's [1988] definition of d-separation for Bayesian networks. We prove that a knowledge base with independence conditions defined by d-separation is a complete spec-ification of a probability distribution.

We present an implemented network generation algorithm that, given an inference problem in the form of a query $Q$ and a set of evidence $E$, generates a network to compute $P(Q|E)$. The generated network is equivalent to a set of ground instances of the sentences in the knowledge base. The algorithm avoids generating many nodes that are irrelevant to the given inference problem, without a separate step to check the relevance of the nodes. Using the semantics defining the probability distribution encoded by a knowledge base, we prove the algorithm correct.

## 2  Representation Language

We would like to represent a class of Bayesian networks using a knowledge base consisting of a collection of probability logic sentences in such a way that a network generated on the basis of the information contained in the knowledge base is isomorphic to a set of ground instances of the sentences. As the formal representation of the knowledge base, we use a subset of Halpern's [1991] probability logic $\mathcal{L}_2^=$. We represent random variables with function symbols and restrict ourselves to using only the equality predicate.

We can represent the information contained in the topology of a Bayesian network, as well as the quantitative information contained in the link matrices, if we can represent all the direct parent/child relations. We express the relation between each random variable and its parents over a class of networks with a collection of universally quantified sentences. The collection of sentences represents the relation between the random variable and its parents for any ground instantiation of the quantified variables. The network fragment consisting of a random variable with function symbol $g$ and its parents with function symbols $f_i$ is represented with a set of sentences of the form

$$\forall X \, P(g(\vec{x}_0) = v_i^0 | f_1(\vec{x}_1) = v_i^1 \wedge \ldots \wedge f_n(\vec{x}_n) = v_i^n) = \alpha,$$

where $X$ is a set of variables and $\vec{x}_0, \ldots, \vec{x}_n$ are subsets of $X$. We have one such sentence for each possible combination of values for the ranges of $g$ and $f_1, \ldots, f_n$. If $g$ has no parents, we use unconditional probability sentences. We also need to express the fact that the values in the range of each function are mutually exclusive and exhaustive. We do so with the following sentences.

$$\forall \vec{x} \, g(\vec{x}) = v_1^0 \vee g(\vec{x}) = v_2^0 \vee \ldots \vee g(\vec{x}) = v_m^0$$

$$\neg (v_j^0 = v_k^0) \text{ for all } j, k \in \{1, \ldots, m\}$$

$$\forall \vec{x} \, f_i(\vec{x}) = v_1^i \vee f_i(\vec{x}) = v_2^i \vee \ldots \vee f_i(\vec{x}) = v_l^i$$

$$\neg (v_j^i = v_k^i) \text{ for all } j, k \in \{1, \ldots, l\}$$

The above probabilistic and logical sentences are all well-formed formulas of the logic $\mathcal{L}_2^=$.

We represent such a collection of sentences with a rule of the form

**Ante:** $f_1(\vec{x}_1) : \{v_1^1, \ldots, v_k^1\} \wedge \ldots \wedge f_n(\vec{x}_n) : \{v_1^n, \ldots, v_l^n\}$
**Conse:** $g(\vec{x}_0) : \{v_1^0, \ldots, v_m^0\}$
**Matrix:** $|g| \cdot |f_1| \cdot \ldots \cdot |f_n|$ entries

where **Matrix** contains the conditional probability of each possible value for $g(\vec{x}_0)$ given each possible combination of values of the $f_i(\vec{x}_i)$; $|f_i|$ is the cardinality of the range of $f_i$; and the universal quantifier is implicit. We call **Matrix** the *link matrix* of the rule. We call the $f_i$ the *antecedents* and $g$ the *consequent*. To identify the antecedents and consequent of a rule $R$, we will use ante$(R)$ for the set of antecedents and conse$(R)$ for the consequent. We will refer to the terms with function symbols, such as $g(\vec{x})$ and $f_1(\vec{x})$, as the *terms of the knowledge base.*

The truth values of the probability sentences are defined with respect to the models of logic $\mathcal{L}_2^=$. A model is a tuple $\langle D, S, \pi, \mu \rangle$, where $D$ is a domain; $S$ is a set of possible worlds; $\pi$ is a function such that for each world $s \in S$, $\pi(s)$ assigns to the predicate and function symbols predicates and functions of the right arity over $D$; and $\mu$ is a discrete probability function on $S$. The semantic value of a formula is defined relative to a model $M$, a possible world $s$, and a value assignment $g$. We interpret a conditional probability sentence of the form $P(A|B) = \alpha$ as shorthand for the sentence $P(A \wedge B) = \alpha \cdot P(B)$. So we have the following semantic definition for each conditional probability sentence represented by our rules, were $g[x/d]$ denotes the assignment of values to variables that is identical to $g$ with the possible exception that element $d$ is assigned to $x$.

$$\llbracket \forall X P(g(\vec{x}_0) = v^0 \mid \wedge_i f_i(\vec{x}_i) = v^i) = \alpha \rrbracket^{M,s,g} = \text{true}$$
if and only if for all $\vec{d} \subseteq D$
$$\mu\{s' : \llbracket g(\vec{x}_0) = v^0 \rrbracket^{M,s',g[X/\vec{d}]} = \text{true and}$$
$$\llbracket f_1(\vec{x}_1) = v^1 \rrbracket^{M,s',g[X/\vec{d}]} = \text{true and} \ldots \text{and}$$
$$\llbracket f_n(\vec{x}_n) = v^n \rrbracket^{M,s',g[X/\vec{d}]} = \text{true}\} =$$
$$\alpha \cdot \mu\{s' : \llbracket f_1(\vec{x}_1) = v^1 \rrbracket^{M,s',g[X/\vec{d}]} = \text{true and} \ldots \text{and}$$
$$\llbracket f_n(\vec{x}_n) = v^n \rrbracket^{M,s',g[X/\vec{d}]} = \text{true}\}$$

This definition says that the universally quantified conditional probability sentence is true if for all ways we can substitute domain elements for the quantified variables, the probability of the consequent and antecedent is equal to $\alpha$ times the probability of the antecedent. The truth values of the nonprobabilistic sentences are defined in the usual way.

We would like a knowledge base to represent a class of Bayesian networks in the sense that a set of ground instances of a subset of the rules is structurally isomorphic to some Bayesian network and contains the same quantitative information.[1] Thus we must have a

---

[1] This does not mean that the knowledge base is isomorphic to a single Bayesian network since each rule can be instantiated multiple times, as is shown in the example in

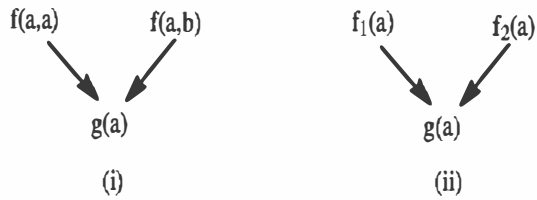

Figure 1: Ramifications of violating the syntactic constraints.

link matrix for each term in the knowledge base. We can guarantee this by requiring every term to be the consequent of some rule:

**C1** (Existence of Link Matrix) Each rule antecedent must also appear as the consequent of some rule.

Furthermore, the link matrix on each rule must be a complete specification of the probabilistic relation between the antecedents and the consequent for all possible ground instantiations of the rules in the knowledge base. To ensure that this property holds, we impose two constraints on the form of the rules. First, to guarantee that the matrix fully specifies the dependencies represented by the individual rule, we require the following:

**C2** (Completeness of Consequent) All variables in the antecedents must appear in the consequent.

Without this constraint we could run into the situation where we have a rule with antecedent $f(x, y)$ and consequent $g(x)$ that gets instantiated twice to produce network fragment (i), shown in figure 1. The matrix associated with this rule is not sufficient to specify the dependency between $g(a)$ and its two direct predecessors $f(a, a)$ and $f(a, b)$ in the generated network.

Second, to guarantee that the matrix fully specifies the dependencies represented by multiple rules, we require that:

**C3** (Uniqueness of Consequents) The knowledge base does not contain two distinct rules that have ground instances with identical consequents.

Without this constraint, we could have a knowledge base that contains one rule with antecedent $f_1(x)$ and consequent $g(x)$ and another rule with antecedent $f_2(x)$ and consequent $g(x)$. These rules could be instantiated to produce network fragment (ii) shown in figure 1. The individual matrices on the two rules are insufficient to specify the dependency between $g(a)$ and its two direct predecessors $f_1(a)$ and $f_2(a)$ in the generated network. If we wish to express that $g(x)$ is influenced by both $f_1(x)$ and $f_2(x)$, this can be repre-sented simply by using the single rule with antecedents $f_1(x)$ and $f_2(x)$ and consequent $g(x)$.

Finally, since a Bayesian network is a directed *acyclic* graph, we must ensure that the knowledge base contains no cycles:

**C4** (Acyclicity) There does not exist a set of ground instances $R_1, R_2, ..., R_n$ of the rules in the knowledge base such that
$\mathsf{conse}(R_1) \in \mathsf{ante}(R_2)$, $\mathsf{conse}(R_2) \in \mathsf{ante}(R_3)$, ..., $\mathsf{conse}(R_n) \in \mathsf{ante}(R_1)$.

These four constraints give us an isomorphic mapping between sets of ground instances of the rules in the knowledge base and Bayesian networks but they also limit the expressiveness of the knowledge base. Constraint C2 restricts the knowledge base from containing recursive rules. For example, we might wish to describe the ancestor relation with a rule that says "if $x$ is the ancestor of $y$ and $y$ is the ancestor of $z$ then $x$ is the ancestor of $z$. Constraint C2 would not allow this since the variable $y$ does not appear in the consequent. Constraint C3 means that we cannot have two rules describing alternative causes of a condition. One might wish to have alternative rules and to choose the most specific one for which information concerning the antecedents is available.

## 3    Bayesian Knowledge Bases

Just as the parent/child relations in a Bayesian network do not in general completely specify a probability distribution, in general neither will a collection of our rules. Using the concept of d-separation, the Bayesian network formalism adds independence information to complete the distribution. We will supplement the probabilistic information contained in our rules with independence information by defining a version of d-separation for knowledge bases. Using an analogue of d-separation to specify independence relations in the knowledge base provides a simple mapping from the semantics of the knowledge base to the semantics of the network. Thus it should be clear to a user what independencies will be encoded in the topology of any network generated from such a knowledge base. With a precise definition of how a knowledge base represents a complete joint probability distribution, we can prove whether or not any network generation algorithm is correct. A generation algorithm is correct if it preserves the distribution when mapping from the knowledge base representation to the network representation.

The concept of d-separation is defined in terms of paths between nodes in a network. So in order to define an analogous notion for a knowledge base, we need to define the concept of a path through the rules in a knowledge base.

**Definition 1** *There is a* **directed edge** *from a*

*ground term $F$ to a ground term $G$ if there is a rule ground instance $R$ such that $F \in \mathsf{ante}(R)$ and $G = \mathsf{conse}(R)$. There is a* **path** *between two ground terms $F$ and $G$ if there is a set of rule ground instances $R$ and a set of ground terms $\mathbf{F} = \{F_1, ..., F_n\}$ appearing in $R$ such that the $F_i$ can be ordered so that there is an edge between $F$ and $F_1$, $F_1$ and $F_2$, ..., $F_n$ and $G$. We call the $F_i$ the ground terms on the path. A ground term $F_i$ on this path is said to have* **converging arrows** *if there are two directed edges on the path pointing to $F_i$.*

*Let $F$ and $G$ be two ground terms. We say that $F$ is a* **direct predecessor** *of $G$ and $G$ is a* **direct successor** *of $F$ if there is a directed edge from $F$ to $G$. We say that $F$ is a* **predecessor** *of $G$ and $G$ a* **successor** *of $F$ if there is a path in the knowledge base between $F$ and $G$ along which all edges point toward $G$. If a ground term has no direct predecessors, it is called a* **root**. *Similarly, if a ground term has no direct successors, it is called a* **leaf**.

Using the definition of path we can now define d-separation for knowledge bases. Notice that our definition of path in a knowledge base results in a definition of d-separation that is almost identical to Pearl's [1988] definition of d-separation in Bayesian networks.

**Definition 2** *If $\mathbf{X}$, $\mathbf{Y}$, and $\mathbf{Z}$ are three disjoint sets of ground instances of terms in a knowledge base KB then $\mathbf{Z}$ is said to* **d-separate** $\mathbf{X}$ *from* $\mathbf{Y}$ *in KB, denoted $\langle X|Z|Y \rangle_D^{KB}$, if along every path between a ground term in $\mathbf{X}$ and a ground term in $\mathbf{Y}$ there is a ground term $W$ satisfying one of the following two conditions: (1) $W$ has converging arrows and none of $W$ or its descendents are in $\mathbf{Z}$, or (2) $W$ does not have converging arrows and $W$ is in $\mathbf{Z}$.*

We now complete the specification of the semantics of the knowledge base by using d-separation to identify probabilistic independencies.

**Definition 3** *Given a set of rules KB of the form specified above and obeying constraints $C1$-$C4$ and given a model $M$ satisfying the sentences corresponding to the rules of KB, KB is called a* **Bayesian knowledge base** *of $M$ if and only if for every three disjoint sets $\mathbf{X}$, $\mathbf{Y}$, and $\mathbf{Z}$ of ground instances of terms in KB if $\langle X|Z|Y \rangle_D^{KB}$ then $\mathbf{X}$ is independent of $\mathbf{Y}$ given $\mathbf{Z}$ in $M$. $\mathbf{X}$ is independent of $\mathbf{Y}$ given $\mathbf{Z}$ in $M$ if the sentences $P(\wedge_i X_i = u_i | \wedge_j Y_j = v_j \wedge_k Z_k = w_k) = P(\wedge_i X_i = u_i | \wedge_k Z_k = w_k)$ over all possible combinations of values $u_i, v_j, w_k$ are satisfied by $M$ for all $X_i \in \mathbf{X}$, $Y_j \in \mathbf{Y}$, and $Z_k \in \mathbf{Z}$.*

With the independencies indicated by d-separation, a knowledge base is now a complete specification of a probability distribution. We will prove this but we first provide a useful lemma.

**Lemma 4** *A ground term in a Bayesian knowledge*

*base is independent of all ground terms which are not its successors, given its direct predecessors.*

*Proof:* Let $KB$ be a Bayesian knowledge base of a probability logic model $M$ and let $G$ and $F$ be any two distinct ground terms contained within any nonempty set of ground instances of the rules of a Bayesian knowledge base KB. We wish to show that given only the direct predecessors of $G$, if $F$ is not a successor of $G$, then $G$ is independent of $F$.

We know that for any path in KB between $G$ and $F$, exactly one of the following must hold:

1. The path is a direct link from $G$ to $F$.
2. The path is a direct link from $F$ to $G$.
3. The path must pass through one of $G$'s direct predecessors.
4. The path must pass through one of $G$'s direct successors.

1) Because $F$ is a successor of $G$, nothing must be shown.

2) Because $F$ is a direct predecessor of $G$, the theorem is trivially true.

3) The path from $F$ to $G$ is blocked by the direct predecessor. So by the definition of d-separation in a knowledge base, the direct predecessor d-separates $G$ and $F$ and by the definition of a Bayesian knowledge base, $G$ is independent of $F$ given the direct predecessor.

4) Since $F$ is not a successor of $G$, for $F$ to be linked to $G$, it must be linked through a successor of $G$. That successor has converging arrows. Such a path is active only if that ground term or one of its successors is given. But we are only given direct predecessors of $G$. Therefore, that path is blocked and $G$ and $F$ are independent. This completes the proof. ☐

**Theorem 5** *A Bayesian knowledge base is a complete specification of a joint probability distribution over the ground terms contained in any non-empty set of ground instances of its rules in which every ground term is the consequent of some rule instance.*

*Proof:* We prove this by showing that the joint probability distribution can be expressed as the product of the conditional probabilities in the link matrices of the rule instances. We define the link matrix of a ground term to be the link matrix of the rule instance for which the ground term is the consequent. Since constraint C3 prohibits the knowledge base from containing two rules that have ground instances with identical consequents, the concept is well-defined.

Let $S$ be any non-empty set of ground instances of the rules in a Bayesian knowledge base $KB$, such that every ground term is the consequent of some rule instance. Let $T = \{\tau_1, ..., \tau_m\}$ be the set of ground terms contained in $S$. We can express a joint probability distribution over $T$ by choosing any ordering of the ground terms and using the chain rule:

$$P(\tau_1, \tau_2, ..., \tau_m) = P(\tau_m \mid \tau_{m-1}, \ldots, \tau_1) \ldots P(\tau_3 \mid \tau_2, \tau_1)$$
$$P(\tau_2 \mid \tau_1) P(\tau_1)$$

We will show that we can choose an ordering of the ground terms to put them in correspondence with the factors of the chain rule expression above such that their link matrices in conjunction with the independencies expressed through d-separation completely specify the conditional probability factors. We will order the ground terms according to the following scheme. We first assign levels to the terms.

- Leaves will be labeled as level zero.

- The level of a ground term is one plus the highest level of any direct successor. (Note that this is well-defined, by the acyclicity of the rules.)

The ground terms are ordered as follows.

- Label the k leaves comprising the bottom level as $\tau_1, \ldots, \tau_k$.

- Label the j ground terms comprising the next level (direct predecessors of the leaves) as $\tau_{k+1}, \ldots, \tau_{k+j}$.

- Label the remaining levels of ground terms in this manner until all are labeled.

Observe that with this ordering, each conditional probability factor $P(\tau_i \mid \tau_{i-1} \ldots \tau_1)$ in the chain rule expression has all of $\tau_i$'s direct predecessors on the right hand side of the conditioning bar and none of $\tau_i$'s successors. So by Lemma 4, each conditional probability factor is completely specified by the link matrix on $\tau_i$. Thus the joint probability distribution is completely specified by the probabilities in the link matrices of the rule ground instances.    □

## 4    Network Generation Algorithm

Our network generation algorithm is query driven. Given a query $Q$ and a set of evidence $E$, we generate a network to compute $P(Q|E)$ such that the probability computed with the network is equal to that defined by the knowledge base semantics. The evidence $E$ will be a conjunction of ground atomic formulas such as $f(a, b) = v$. Since probability logic can represent a large range of possible queries, we might ask what kinds of queries can be answered using the information in a Bayesian knowledge base. If we wish to infer a point probability for our query, then the semantics of our rules requires that the query be a ground formula. A non-ground query would be either universally or existentially quantified, or a combination of both. But it is clear from the semantic definition of the universally quantified conditional probability sentences that we can only infer precise probabilities of individual elements of the domain and thus only of ground formulas. For universally quantified formulas, we can only infer upper bounds and for existentially quantified formulas, we can only infer lower bounds. So our algorithm takes a query $Q$ in the form of a ground term (e.g. $g(a)$), a collection of evidence $E$ in the form of a set of ground atomic formulas, and a Bayesian knowledge base $KB$, and generates a network to compute $P(Q|E)$. The probability computed with the generated network is equal to that defined by the knowledge base semantics.

The key idea behind the algorithm is that since the rules in the knowledge base are structurally similar to Horn-clauses, we can use a backward-chaining theorem prover to search through the rules for paths between the evidence and the query. The generated network is just the resulting proof tree. By simply backward chaining on the query and on the evidence formulas the algorithm generates all relevant nodes and avoids generating barren nodes [Shachter, 1988], which are nodes below the query that have no evidence nodes below them. Such nodes are irrelevant to the computation of $P(Q|E)$. The algorithm has been implemented in Lisp.[2]

### Network Generation Algorithm

> **Input:** A Bayesian knowledge base, a query in the form of a ground term, and evidence in the form of a set of ground atomic formulas.
> **Output:** A network to compute the probability of the query given the evidence.

> #### Procedure:
> 1. Backward chain on the query, generating all its predecessors. Call this the query's predecessor network.
> 2. For each evidence formula without a corresponding node in the query's predecessor network do:
>    - Backward chain on the evidence formula. If an antecedent has a corresponding node in the network generated so far, create a link to it. Otherwise create a new node.

Since the query and the evidence formulas are ground and all variables in the antecedents of a rule must appear in the consequent (constraint C2), the nodes generated by backward chaining will always be ground terms. So the generated network is a subset of the set of ground instances of the rules in the knowledge base, as desired. Furthermore, since the algorithm backward chains until a root term is reached, the roots in the network correspond to roots in the knowledge base.

---

[2]We have implemented and proven correct a slightly more complex algorithm that is guaranteed to generate the smallest network necessary to answer a given query but do not discuss it here due to space limitations. The implementation interfaces to the IDEAL [Srinivas and Breese, 1990] inference system and is available via anonymous ftp to ftp.cs.uwm.edu in pub/tech_reports/ai/BNG.tar.Z.

| R1 | **Ante:** Neighborhood(x):{b,a,g}<br>**Conse:** Burglary(x):{+,-}<br>**Matrix:** (6 entries) | R2 | **Ante:** Burglary(x):{+,-}, Quake:{t,m,s}<br>**Conse:** Alarm(x):{+,-}<br>**Matrix:** (12 entries) |
|---|---|---|---|
| R3 | **Ante:** Quake:{t,m,s}<br>**Conse:** Radio:{+,-}<br>**Matrix:** (6 entries) | R4 | **Ante:** Alarm(x):{+,-},<br>Neighbor(n,x):{+,-}<br>**Conse:** Phone-call(n,x):{+,-}<br>**Matrix:** (8 entries) |
| R5 | **Ante:** Report(x):{+,-},<br>Burglary(x):{+,-}<br>**Conse:** Recovered(x):{+,-}<br>**Matrix:** (8 entries) | R6 | **Ante:**<br>**Conse:** Neighbor(n,x):{+,-}<br>**Matrix:** (prior) |
| R7 | **Ante:**<br>**Conse:** Quake:{t,m,s}<br>**Matrix:** (prior) | R8 | **Ante:**<br>**Conse:** Neighborhood(x):{b,a,g}<br>**Matrix:** (prior) |
| R9 | **Ante:**<br>**Conse:** Report:{+,-}<br>**Matrix:** (prior) | | |

Figure 2: Example Bayesian Knowledge Base.

We can view the network generation algorithm as mapping from the knowledge base representation of a probability distribution to a Bayesian network representation of that distribution. We now prove that the algorithm is correct in the sense that the mapping preserves the probability distribution specified by the semantics of the knowledge base.

**Theorem 6** *The Bayesian network generation algorithm is correct in the following sense. Let $N$ be a Bayesian network generated from a Bayesian knowledge base $KB$ in response to a query $Q$ and evidence $E$. Let $P_N$ be the probability distribution over the ground terms in $N$, as specified by $N$. Let $P_{KB}$ be the probability distribution over the ground terms in $N$, as specified by $KB$. Then $P_N$ and $P_{KB}$ are identical. Thus $P_{KB}(Q|E) = P_N(Q|E)$.*

*Proof:* By theorem 5, the Bayesian knowledge base from which the network is generated is a complete specification of the joint distribution over the ground terms in the rule instances representing the generated network. Since the proof of theorem 5, can be duplicated for Bayesian networks using the network definition of d-separation, the distribution over the ground terms as specified by the network and by the knowledge base are equal.

## 5    Example

Consider the knowledge base describing a burglary domain shown in figure 2.[3] The rules express the following information.

**R1:** The type of neighborhood someone lives in influences whether their house will be burglarized.

**R2:** Both a burglary and an earthquake can cause someone's alarm to go off.

**R3:** An earthquake is often reported on the radio.

**R4:** If someone's alarm goes off, his neighbor is likely to call him.

**R5:** Whether or not someone's house was burglarized and whether they filed a report influences whether the stolen goods will be recovered.

**R6–R9:** These rules specify unconditional probabilities for the root terms.

All random variables are binary-valued except Quake which has values {tremor, moderate, severe} and Neighborhood which has the values {bad, average, good}.

Suppose we have the evidence

{Radio:=+,   Neighbor(Watson,Holmes)=+,
Phone-call(Watson,Holmes)=+,
Neighbor(Moriarty,Holmes)=+,
Phone-call(Moriarty,Holmes)=+}

---

[3]This example is based on one presented by Breese [1992].

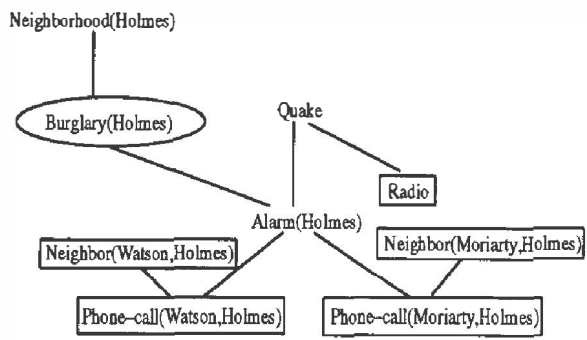

Figure 3: Network generated for query Burglary(Holmes).

and we wish to know the posterior probability of Burglary(Holmes). The network generation algorithm will generate the network shown in figure 3. The query node is indicated by an ellipse and evidence nodes by boxes. Notice that rule R4 is used twice: once for Watson and once for Moriarty. This illustrates how the use of quantification allows us to represent a potentially large class of networks. Notice also that we do not generate the nodes Report(Holmes) or Recovered(Holmes) since they are irrelevant to this particular inference problem.

## 6    Related Work

In some early work, Poole [1991] and Poole and Horsch [1990] present a method for representing Bayesian networks with Horn-clauses. This is considered to be a template representation and is not provided with a logical semantics. Poole [1993] presents a framework for logic-based abduction that incorporates probabilities. He describes a logical language for specifying a probabilistic Horn abduction theory that consists of definite clauses and disjoint declarations. A disjoint declaration specifies the probabilities on a set of mutually exclusive hypotheses. His probabilistic Horn abduction language can be viewed as a logic of discrete Bayesian networks. He shows how to translate between Bayesian networks and probabilistic Horn abduction theories and proves that the probability distribution represented by a Bayesian network is equivalent to that represented by its probabilistic Horn abduction theory translation. He provides a model-theoretic semantics for his language in terms of Bacchus's propositional probability logic [Bacchus, 1990].

Goldman and Charniak [1990, 1993] present a language, Frail3, for representing belief networks and outline their associated generation algorithm. The language includes numerous generic causal models, such as the noisy-OR gate [Pearl, 1988], for specifying link matrices when complete probabilistic information is not available. Networks are generated by a forward and backward chaining TMS type system. Frail3 represents network dependencies by rules with variables, but the semantics of the variables is not specified.

Breese [1992] presents a language, Alterid, that can represent the class of Bayesian networks and influence diagrams with discrete-valued nodes. Probabilistic relations are specified with universally quantified conditional probability sentences. Breese presents a detailed algorithm to generate networks from an Alterid knowledge base. Given a query $Q$ and a set of evidence $E$, the algorithm uses both forward and backward chaining to generate a network to compute $P(Q|E)$.

Breese does not constrain the syntax of the knowledge base as we do. Thus a set of ground instances of the rules may not represent a single network. In this case, the generation algorithm must choose between networks to generate. Breese does not provide a formal semantics for his representation and does not prove his algorithm correct in the sense that we do. By assuming that the generated network is what Breese calls a Bayesian interpretation network, he shows that the algorithm generates all nodes necessary to compute $P(Q|E)$. A Bayesian interpretation network is essentially an I-map of a probability distribution, but not necessarily a minimal I-map. But he does not prove that the algorithm generates a network to correctly compute $P(Q|E)$.

Bacchus [Bacchus, 1993] sketches a framework for knowledge-based construction of Bayesian networks using first-order statistical probability logic. The language is first-order logic augmented with the ability to express assertions about proportions. He discusses representing a class of networks with a set of universally quantified conditional probability sentences, each of which expresses a piece of local statistical information. Because he is interested in representing probabilistic information of varying degrees of specificity he does not constrain his representation in the ways that we do. But because he does not impose constraints, he cannot show that a knowledge base consisting of separate pieces of statistical information can completely specify either a Bayesian network or a joint probability distribution. He suggests generating a network by looking for chains of influence and instantiating the sentences but does not provide a network generation algorithm.

## 7    Current and Future Research

One of the objectives of the present work is to introduce quantification into the Bayesian network formalism. Here we have explored only universal quantification outside the scope of the probability operator. Since the logic $\mathcal{L}_P^=$ permits arbitrary quantification, we could think of extending the current subset of probability logic to include existential quantification, as well as quantification within the scope of the probability operator. For example, we might wish to write a sentence like $\forall y \; P(\exists x \; Q(x)|R(y)) = 0.7$. To extend the representation in this way, we need to identify particular sentence forms that can be used to generate networks, as we have done for sentences with a universal quantifier outside the scope if the probability operator.

One application to which the technique of dynamic network generation seems particularly well suited is the representation of time, since with temporal problems we typically do not know a priori the exact time points at which we will have or want to compute information. To formulate an algorithm to generate temporal Bayesian networks, we first require a representation for the knowledge base that can express information about temporal ordering as well as probabilities of temporal objects. Haddawy [Haddawy, 1991, Haddawy, 1994] presents a logic that integrates both time and probability and hence has the needed expressiveness.

### Acknowledgements

I would like to thank Kate Fowler, Bob Krieger and Meliani Suwandi for many helpful discussions and for implementing the generation algorithm, and Susan McRoy for helpful comments on an earlier draft. This work was partially supported by NSF grant #IRI-9207262.